\def\year{2022}\relax
\documentclass[letterpaper]{article} 
\usepackage{aaai22}  
\usepackage{times}  
\usepackage{helvet}  
\usepackage{courier}  
\usepackage[hyphens]{url}  
\usepackage{graphicx} 
\usepackage{natbib}  
\usepackage{caption} 
\DeclareCaptionStyle{ruled}{labelfont=normalfont,labelsep=colon,strut=off} 
\nocopyright
\title{Learning Context-Aware Embedding for Person Search}
\author{
    Shihui Chen\textsuperscript{\rm 1}\thanks{This work is done when Shihui Chen is an intern at MEGVII Technology.}, 
    Yueqing Zhuang\textsuperscript{\rm 2}, Boxun Li\textsuperscript{\rm 2}
}   
\affiliations{
    \textsuperscript{\rm 1} Beihang University, 
    \textsuperscript{\rm 2} MEGVII Technology\\
    
    shihuichen@buaa.edu.cn, zhuangyueqing@megvii.com,
    liboxun@megvii.com

}

\usepackage{booktabs}
\usepackage{multirow}
\usepackage{amsmath}
\usepackage{amsfonts}
\usepackage{subfigure}

\begin{document}

\maketitle

\begin{abstract}
    Person Search is a relevant task that aims to jointly solve Person Detection and Person Re-identification (re-ID). Though most previous methods focus on learning robust individual features for retrieval, it’s still hard to distinguish confusing persons because of illumination, large pose variance, and occlusion. Contextual information is practically available in person search task which benefits searching in terms of reducing confusion. To this end, we present a novel contextual feature head named Attention Context-Aware Embedding(ACAE) which enhances contextual information. ACAE repeatedly reviews the person features within and across images to find similar pedestrian patterns, allowing it to implicitly learn to find possible co-travelers and efficiently model contextual relevant instances' relations.  Moreover, we propose Image Memory Bank to improve the training efficiency. Experimentally, ACAE shows extensive promotion when built on different one-step methods. Our overall methods achieve state-of-the-art results compared with previous one-step methods.
\end{abstract}

\section{Introduction}

    Most visual surveillance systems consist of a person detector locating persons and a person re-identification (re-ID) to recognize the identities of persons. The system is usually divided into two independent tasks: Pedestrian Detection and Person Re-identification (re-ID). Based on real applications, person search~\cite{commonness, joint} task is proposed to handle this situation, which jointly locates and retrieves a query person from a gallery of scene images captured by different cameras. Furthermore, person search benefits in terms of convenience and efficiency, drawing much attention in practical applications.
    
    Traditional pedestrian detection focuses on localizing persons precisely while re-ID pays more attention to searching persons from different cameras. Standard methods address person search in a two-step manner, cascading a pedestrian detector with a re-ID feature extractor and training them independently~\cite{PRW, multi-scale-matching, Mask-Guided-Two-Stream, ReID-Driven, TCTS}. In detail, single pedestrian images are cropped from the scene images according to the boxes produced by a detector and fed into a standard person re-ID model for identification. In contrast, another series of methods called one-step methods optimize these two tasks in a unified framework~\cite{joint}, localizing and extracting features of detected persons at the same time.
    
    \begin{figure}[t]
        \centering
        \centerline{\includegraphics[width=0.9\columnwidth]{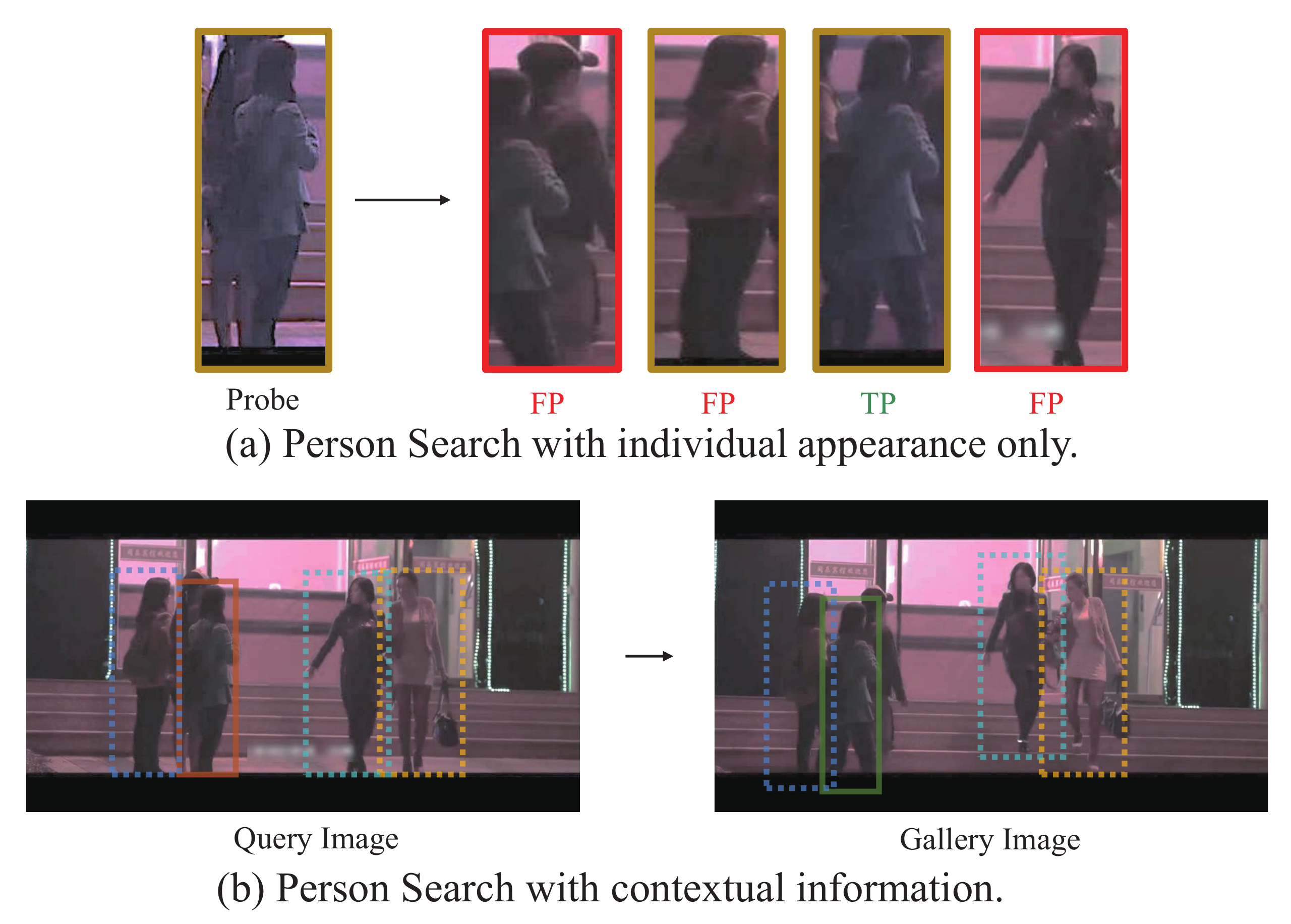}}
        \caption{Importance of searching with contextual information. With contextual information around, the searching becomes much easier than with confusing individual appearance. Boxes in the same color represent the same person. 'FP' represents False Positive, while 'TP' represents True Positive.}
        \label{fig:context-aware-motivation}
    \end{figure}
    
    Although significant progress has been made in one-step~\cite{joint, IAN, NPSM, CTXGraph, NAE, decoupled} and two-step person search~\cite{Mask-Guided-Two-Stream, IGPN, ReID-Driven, TCTS, RCAA}, it's still hard to match correct persons by individual box-level re-ID features. Different person re-ID task, person search is given scene images as input, making it available for the relationship of persons in one image. In reality, people are likely to walk in groups and communicate with each other. Even when people are walking alone, other neighboring pedestrians in the same scene also provide important contextual clues for searching. For example, in Figure~\ref{fig:context-aware-motivation}, it's hard to distinguish the right person by appearance only. However, this turns to be easier if we resort to the contextual persons. This indicates the importance of contextual clues when individual features are uncertain because of illumination, large pose variance, occlusion, etc.
    
    In this paper, we aim at utilizing contextual clues in searching but avoiding the hand-craft process to find possible co-travelers in previous methods, which may introduce noises. To this end, a lightweight plug-in module, named Attention Context-Aware Embedding(ACAE) head, is proposed, inspired by Graph Attention Networks(GATs)\cite{velivckovic2018GAT}. ACAE tries to aggregate contextual information from individual features within and across images. Its attention-based aggregation assigns higher weights to persons who are more likely to become co-travelers and lower weights to those less possible, which corresponds to an implicit and learnable co-travelers discovery process. Training with ACAE requires paired images, which doubles the computation to extract features. To improve the efficiency of end-to-end training with ACAE, we further introduce the Image Memory Bank(IMB) strategy to avoid duplicate image features extraction.
    
    ACAE head is easy to be incorporated into different one-step methods as long as we place it after the module which extracts individual appearance features and apply the IMB strategy. Full experiments show that ACAE could further promote the performance of different one-step methods. In the end, together with a new pipeline with a Transformer-based detector, our overall methods reach 93.9/46.2 mAP on CUHK-SYSU and PRW datasets respectively.

    Our contributions can be summarized in three-folds:

    \begin{itemize}
        \item We propose a novel feature head for learning contextual clues between two images in person search, named Attention Context-Aware Embedding(ACAE) head. ACAE aggregates features of persons within and across images and learns to identify possible co-travelers with attention mechanism in end-to-end training.
        \item To improve the training efficiency, we propose the training strategy with Image Memory Bank, which introduces little additional computation but makes it efficient to train with pair data.
        \item Our ACAE head is easy to train and could be incorporated into most previous one-step methods to achieve better performance. Our overall pipeline built on Transformer achieves state-of-the-art performance.
    \end{itemize}

\section{Related Work}

    \subsection{Person Search}  

    Person Search has attracted wide attraction from the computer vision community for its application. Generally, person search could be divided into detection and re-identification. Therefore, a series of \textit{two-step} methods~\cite{PRW, multi-scale-matching, Mask-Guided-Two-Stream, ReID-Driven, TCTS} are proposed. Among them, \cite{PRW} makes a thorough investigation on the combination of different detectors and re-ID models. \cite{multi-scale-matching} analyses the bottleneck of person search and addresses the multi-scale challenge by Cross-Level Semantic Alignment (CLSA) Loss. \cite{Mask-Guided-Two-Stream} indicates that separating the two tasks yields better overall performance. To avoid noisy information from the background, they build another CNN stream for the re-ID phase, which extracts additional foreground person features. \cite{ReID-Driven} develops a differentiable ROI Transform layer to refine localization by re-ID loss. \cite{TCTS} points out the inconsistent problem between the two tasks, and reduces the number of distracting boxes by incorporating the query boxes in detection. For re-ID training, they provide a dataset mixed with detection boxes and dynamically mine hard samples in terms of quality and hardness.
    
    Another series of end-to-end methods called \textit{one-step} methods~\cite{joint, IAN, HOIM, NAE, decoupled} combine the two tasks into a unified framework with a shared feature extractor. One-step methods have also attracted much attention due to their better efficiency and comparable performance. The framework of the first one-step model~\cite{joint} adopts Faster-RCNN~\cite{faster-rcnn} in detection, with an additional feature head for re-ID. During training, they also propose Online Instance Matching (OIM) loss, which is much more efficient than the cross-entropy loss to train considering the sparse identities of each person in mini-batch. Successively, improved loss functions like Center Loss~\cite{IAN}, HOIM Loss~\cite{HOIM} are proposed to better supervise re-ID. Moreover, \cite{NAE} reconciles the contradiction between classification and retrieval by proposing Norm-Aware Embedding (NAE) head, which decouples the embedding features into the norm and angle space for detection and re-ID respectively. In \cite{decoupled}, a novel person search pipeline is proposed with one-stage detectors. Besides, memory-reinforced mechanism is used to boost feature learning, alleviating limited samples problem in mini-batch. Apart from these, there are other novel frameworks combining reinforcement learning~\cite{RCAA}, or introducing convolutional Long Short-Term Memory (LSTM) network to refine localization~\cite{NPSM}.
 
    \begin{figure*}[thb]
        \centering
        \includegraphics[width=\textwidth]{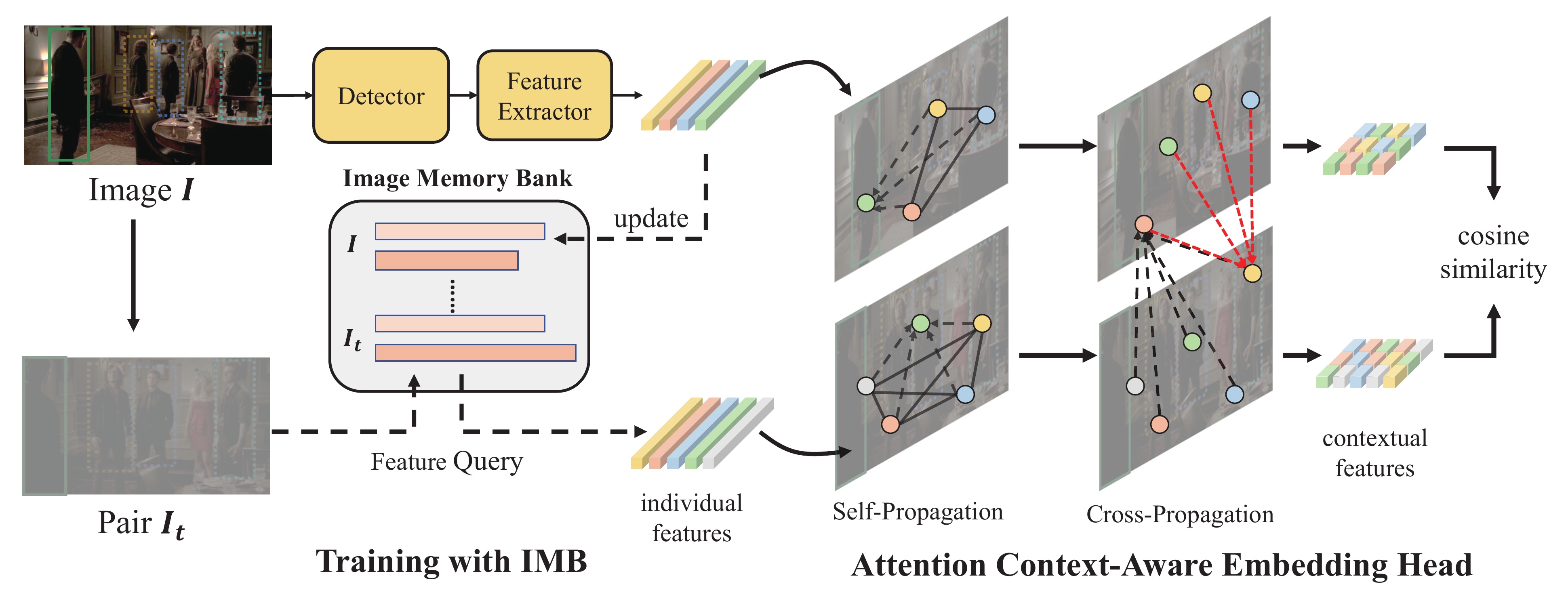}
        \caption{An overview of our methods. Our proposed pipeline consists of a detector and a feature extractor for locating people and extracting the corresponding features. Attention Context-Aware Embedding head follows the appearance feature module to extract contextual features. Then during training, Image Memory Bank is adopted for improving training efficiency.}
        \label{fig:overall-pipeline}
    \end{figure*}
    
    \subsection{Person search with Contextual Clue}
    
    Although previous methods make great progress by promoting the discrimination of individual features, few works pay much attention to contextual information. Individual features usually suffer from scale variation, occlusion, or illumination, while contextual information could be the additional robust clues. To utilize information from contextual persons, one may divide it into two stages: matching stage to find possible co-travelers between two images, and refining stage to adjust the target similarity with these found co-travelers. This two stages pattern could be clearly found in previous related methods. Concretely, CTXGraph\cite{CTXGraph} uses part-based features with a manual similarity threshold to find possible co-travelers, and GCN for the refining stage to judge if the target pair is of the same identity. However, manual matching methods to find co-travelers could introduce unnecessary noise, thus misguiding the refining stage. Differently, our proposed ACAE is designed to combine the two stages together by extracting contextual features for contextual similarity. The matching stage is implicitly embedded by the attention module which assigns larger weights for persons around who are more likely to become co-travelers.

\section{Methods}

    Traditional one-step methods consist of a backbone detector and a feature extraction branch. We follow the same paradigm to build our baseline model but take a little step forward to introduce the transformer-based detector into our framework, to explore its ability in person search task. Specifically, we adopt the off-the-shelf Deformbale-DETR\cite{deformable-detr} model for detection. While for re-ID embedding, given produced proposals, features extracted from ${C_4}$ and ${C_5}$ stage of a ResNet50 backbone through ROIAlign~\cite{mask-rcnn} would be reduced to 128 dimensions by linear layers respectively and then concatenated together as the final embeddings of persons.
    
    In this section, we will present our Attention Context-Aware Embedding head in detail and describe the training strategy with Image Memory Bank. The overall pipeline is shown in Figure~\ref{fig:overall-pipeline}.

    \subsection{Attention Context-Aware Embedding Head}

    \textbf{Motivation:} Using contextual clues for matching confusing persons is quite a natural behavior for human performance. In brief, contextual features could be defined as similar persons' patterns between two images. In feature-learning view, that should be the weighted aggregation of selective neighbor persons' features, of which the weights should be related to their possibility of becoming co-travelers. Inspired by Graph Attention Networks\cite{velivckovic2018GAT} for extracting useful representation for graph data based on nodes' similarity, an attention graph model is introduced.
    
    \noindent \textbf{Formulation:}
    To decide whether a person around target could become co-traveler, one must discern if it also appears in the gallery image, so a pair of input is required. Consider two images ${A, B}$, we take a pair of appearance features sets ${ \left\{\mathbf{f}^{A}_{i}\right\}_{i=1}^{m}}$, ${\left\{\mathbf{f}^{B}_{j}\right\}_{j=1}^{n}}$ as input, which could be easily extracted by any person search models. Then complete undirected graphs ${\mathcal{G}_{A}, \mathcal{G}_{B}}$ are built separately, with each node ${\mathbf{p}_i, \mathbf{q}_j }$ initialized with individual features ${\mathbf{f}^{A}_{i}, \mathbf{f}^{B}_{i}}$. 
    
    Take the feature aggregation in ${\mathcal{G}_{A}}$ as an example. The intra-image feature ${\bar{\mathbf{p}}_{i}}$ of node ${i}$ is aggregated in attention way as follows:
    
    \begin{equation}
        \begin{aligned}
            e_{i,j} &= f_{Q}(\mathbf{p}_i)^{T}f_{K}(\mathbf{p}_j) \\
            w_{i,j} &= \frac{\exp(e_{i,j})}{\sum_{j} \exp(e_{i,j}) } \\
            \bar{\mathbf{p}}_{i} &= \operatorname{LN}\left( \mathbf{p}_{i} + \sum_{j} w_{i, j} f_{V}(\mathbf{p}_j) \right) \\
        \end{aligned}
    \end{equation}
    
    where ${w_{i,j}}$ is the normalized weight of neighbor node ${j}$ for node ${i}$. ${f_Q, f_K, f_V}$ are the linear projections for the query, key, value input. Appearance feature ${\mathbf{p}_i}$ and its aggregated neighbors, normalized by layer normalization(LN)~\cite{layernorm}, constitute the overall intra-image representation for each node.

    The intra-image feature ${\bar{\mathbf{p}}_i}$ of node ${i}$ is used as the query to find possible co-travelers in gallery image ${B}$. Concretely, the attention mechanism is applied likewise, but with the neighbors of node ${i}$ in ${\mathcal{G}_{A}}$ set to all nodes in ${\mathcal{G}_{B}}$ so that each possible pair could be compared. This results in the inter-image feature ${\hat{\mathbf{p}}_i}$ of node ${i}$ computed in the following way:
    
    \begin{equation}
        \begin{aligned}
            e^{\prime}_{i,j} &= f_{Q}(\bar{\mathbf{p}}_i)^{T}f_{K}(\mathbf{q}_j) \\
            w^{\prime}_{i,j} &= \frac{\exp e^{\prime}_{i,j}}{\sum_{j} \exp e^{\prime}_{i,j} } \\
            \hat{\mathbf{p}}_{i} &= \operatorname{LN}\left( \bar{\mathbf{p}}_{i} + \sum_{j} w^{\prime}_{i, j} f_{V}(\mathbf{q}_j) \right) \\
        \end{aligned}
    \end{equation}
    
    Apparently, ${\hat{\mathbf{p}}_i}$ is the aggregation of helpful persons' features which we desire. To strengthen its representation ability, we apply another linear transform, resulting in the final contextual representation ${\tilde{\mathbf{p}}_i}$:

    \begin{equation}
        \begin{aligned}
            \tilde{\mathbf{p}}_i &= \operatorname{LN}\left(\hat{\mathbf{p}}_i + \text{MLP}\left(\hat{\mathbf{p}}_i \right) \right) \\
        \end{aligned}
    \end{equation}

\begin{figure}[t]
        \centering
        \includegraphics[width=0.6\columnwidth]{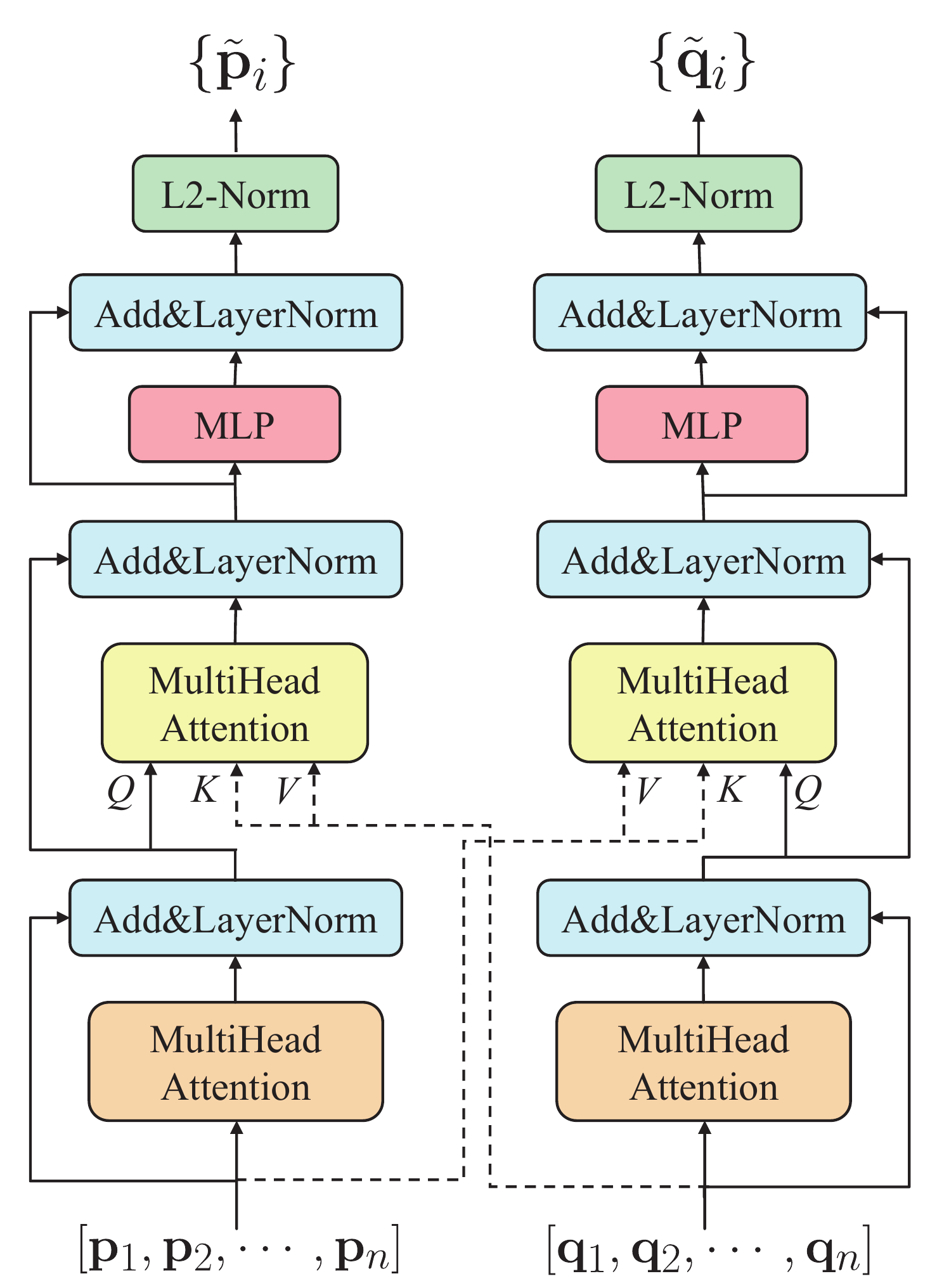}
        \caption{Detailed structure of our proposed Attention Context-Aware Embedding head.}
        \label{fig:acae_detail_structure}
    \end{figure}

    Symmetric feature aggregation is performed for features in graph ${\mathcal{G}_{B}}$ simultaneously. In practice, we extend the attention module to multi-head attention\cite{transformer} to strengthen representation ability. This results in our overall Attention Context-Aware Embedding(ACAE) head, as Figure~\ref{fig:acae_detail_structure} shows. ACAE is quite similar to the decoder structure in Transformer since it fits our design targets well. However, ACAE differs in the settings of K, Q, V, which is essential to make ACAE work.
    
    During inference, for a pair of node ${(i, j)}$ in graph ${\mathcal{G}_A, \mathcal{G}_B}$ respectively, the corresponding contextual similarity ${s^{c}_{i,j}}$ is based on all features extracted:

    \begin{equation}
        s^{c}_{i,j} = \frac{1}{3} \left(
            \bar{\mathbf{p}}_{i}^{T} \bar{\mathbf{q}}_{j}
            + \hat{\mathbf{p}}_{i}^{T} \hat{\mathbf{q}}_{j}
            + \tilde{\mathbf{p}}_i^{T} \tilde{\mathbf{q}}_{j}
        \right)
    \end{equation}
    
    Using final contextual features ${\tilde{\mathbf{p}}_i}$ only could largely boost searching performance, but with intra-image and inter-image features, the performance is further enhanced since they help balance the noises, as our ablation study would show. Then overall similarity for searching is a weighted sum of individual and contextual similarity:

    \begin{equation}
        s_{i,j} = \lambda s^{c}_{i,j} 
        + (1 - \lambda) \mathbf{p}_{i}^{T} \mathbf{q}_{j} \\
    \end{equation}
    
    where ${\lambda}$ is a hyper-parameter. Setting ${\lambda}$ to 0.4 yields the best promotion in practice. Moreover, in one gallery image, no more than one candidate person could be matched to the query target. Thus, we take a simple strategy to rescale all candidates' scores in the same gallery image ${G}$, to reduce disturbance from useless candidates in matching. Then we get the final similarity ${s^{r}_{i,j}}$ for matching:
    
    \begin{equation}
        \begin{aligned}
            c_{i,j} &= \frac{\exp(s_{i,j})}{\sum_{j \in G} \exp(s_{i,j})} \\
            s^{r}_{i,j} &= \frac{c_{i,j}}{\max_j c_{i,j}} s_{i,j} \\
        \end{aligned}
    \end{equation}

    ACAE features its intra- and inter-image attention mechanism to assign distinct weights to persons around to form helpful context. It costs little additional computation but is proved to be effective in our experiments. Moreover, ACAE only relies on a pair of appearance features as input, which makes it easy to extend after the feature branch of any other person search models. Our experiments with different one-step models further prove its extensive effectiveness on various models.

    \subsection{Context-Aware Image Memory Bank.}

    As illustrated before, training with ACAE requires pair images as input and as many ground-truth matched person pairs as possible between two images as the positive samples. So before training, to every image ${I}$, we appoint a pair image ${I_t}$ which shares the most overlapped labeled person identities with ${I}$. However, this requires a two times larger batch size and computation for extracting individual features. To overcome the low efficiency of training, we further introduce the Image Memory Bank(IMB) strategy.
    
    Specifically, in IMB, we maintain features at the image level. For every image ${I}$, we use a look-up table ${L_{I} \in \mathbb{R}^{K_l \times d}}$ to store the features of labeled persons, and a list ${U_I \in \mathbb{R}^{K_u \times d}}$ to store features of unlabeled persons, where ${d}$ is the feature dimension and ${K_l, K_u}$ represents the ground-truth number of labeled and unlabeled persons in image ${I}$. During training, we take out all features in ${L_{I_t}, U_{I_t}}$ of pair image ${I_t}$ to form the paired input for ACAE, leaving the former training process unchanged. Simultaneously real-time extracted labeled and unlabeled persons' features ${\bar{v}_{I}, \bar{u}_{I}}$ of ${I}$ from preceding modules are used to update features of ${I}$ in IMB in the following way:
    
    \begin{equation}
        \left\{
            \begin{aligned}
            v_{I} &\gets \bar{v}_{I} \quad \forall v_{I} \in L_{I} \\
            U_{I} &\gets \left\{\bar{u}_{I} \right\} \\
            \end{aligned}
        \right.
    \end{equation}
    
    IMB only introduces little computation to update features and hardly changes the primary training process, which also makes it easy to incorporate ACAE into training pipeline of other person search models. Our further experiments show its efficiency in training.
    
\section{Experiments}

    In this section, we conduct thorough ablation experiments and visualization to illustrate the effectiveness of our proposed methods on CUHK-SYSU dataset. Then we compare the overall performance of our methods with state-of-the-art methods.

    \subsection{Datasets and Settings}

    \textbf{CUHK-SYSU dataset.} One commonly accepted dataset in person search task is CUHK-SYSU~\cite{joint}. It contains people images from two sources: street snaps and movies, consisting of 18,184 images and 8,432 query persons in total. The dataset is partitioned into two parts without overlap. The trainset contains 11,206 images and 5,532 labeled persons, and the test set includes 6,978 images and 2,900 query persons. Following the same settings with previous methods, our methods are evaluated when gallery size is set to 100 if not specified. 

    \textbf{PRW dataset.} PRW~\cite{PRW} is another large-scale person search dataset containing 11,816 video frames captured by six cameras. The labeled identities range from 1 to 932, while 482 of them are used for training and the other 2057 images and 450 identities for testing.

    \textbf{Evaluation Protocols.} We follow the same evaluation protocols with previous methods~\cite{joint, NAE}. To evaluate pedestrian detection performance, we choose the common Average Precision(AP) and recall rate. While evaluating re-ID performance, we calculate Cumulative Matching Curve(CMC) and Mean Average Precision(mAP). But unlike re-ID task, a person in the gallery matches the query identity not only if their features are most similar but also with IoU between detected boxes and ground truth boxes greater than 0.5. Then we calculate the AP for each query identity and average them to get mAP.

    \subsection{Implementation Details}
    
    As mentioned before, we take the off-the-shelf Deformbale-DETR as the detector in our one-step framework and keep most of the settings the same as default ones without bells and whistles. As for the re-ID embedding part, we also follow the same configuration with previous methods for most settings. A ResNet50 pre-trained from ImageNet is used as our feature extractor. To train the model, we use the Adam optimizer with betas set to ${(0.9, 0.999)}$ and weight decay rate set to ${1 \times 10^{-4}}$. For training the ACAE head, besides IMB, we freeze the loss of ACAE but update IMB in the first epoch to stabilize the features in IMB and avoid interfering with the training of the main branch. OIM loss is employed to train the ACAE head, with its loss weight set to 0.1 in all experiments. All models in this paper are trained for 30 epochs in total with multi-scale augmentation, the training size is randomly chosen from 480 to 900 and the max length of an image is set to $1500$. During training, we take the simple step decay strategy to adjust the learning rate. The initial learning rate is set to ${2 \times 10^{-4}}$, then decays by 0.1 at epoch 15 and 25.
    In inference, we follow the same settings as previous methods to ensure the fairness of comparison. Images are resized to have at least 1500 pixels on the long side and 900 pixels on the short side. Furthermore, we set the nms threshold to 0.4 and filter out boxes with confidence under 0.5. Our implementation is based on the Pytorch framework, and the network is trained on Nvidia V100 with batch size set to 4.
    
    \subsection{Analytical Experiments}
    
    In this section, we explore full experiments to analyze the effectiveness of our proposed methods. To the beginning, we conduct an ablation study to show how different settings of ACAE would affect its performance. Then we evaluate the promotion of ACAE when built on various methods and compare it with related re-ranking methods. Finally, we visually inspect the learned contextual information of ACAE to illustrate its matching performance.
    
    \begin{table}[tbp]
        \begin{center}
        \setlength{\tabcolsep}{1.5mm}{
        \begin{tabular}{lcllll}
        \toprule
        \textbf{Settings} & \textbf{batch} & \textbf{mAP} & \textbf{Top-1} & \textbf{Top-5} & \textbf{Top-10} \\ \midrule
        baseline  & 2          & 91.90 & 92.52 & 96.93 & 97.93 \\
        baseline  & 4          & 92.59 & 93.31 & 97.76 & 98.31 \\
        +ACAE     & ${2 \times 2}$          & 92.92 & 93.79 & 97.72 & 98.28 \\
        +ACAE/IMB & 4          & 93.87 & 94.69 & 98.10 & 98.59 \\ \bottomrule
        \end{tabular}
        }
        \end{center}
        \caption{Ablation results of ACAE and IMB strategy.}
        \label{table:ablation-study}
    \end{table}
    
    \textbf{Influence of different ACAE settings.} We present the ablation results of the introduced ACAE and IMB strategy in Table~\ref{table:ablation-study}. As stated before, training ACAE without IMB requires double data in one batch. Comparing to the baseline model with batch size set to 2, ACAE is able to achieve a 1.0/1.2 performance gain. After applying IMB, training rids the double data constraint and benefits from the larger real batch size, leading to another 0.9 mAP promotion. It's also noteworthy that without IMB, overall training would cost about 37h, while this would be reduced to around 20h with IMB, slightly slower than baseline, which costs 16h. These all prove the efficiency and efficacy of the IMB strategy.
    
    We further analyze how different settings of features used to compute similarity affect the overall performance in Table~\ref{table:feature-ablation}. Clearly, using final contextual features only could bring much promotion from baseline. Introducing intra-image and inter-image features further boost the overall performance since they help balance noises in the final features.

    \begin{table}[h]
        \begin{center}
        \setlength{\tabcolsep}{1.4mm}{
        \begin{tabular}{@{}l|ccc|cc@{}}
        \toprule
        configuration  & intra ${\bar{\textbf{p}}}$ & inter ${\hat{\textbf{p}}}$ & final ${\tilde{\textbf{p}}}$ & mAP & Top1 \\ \midrule
        baseline       &   &       &       & 92.59 & 93.31 \\
        intra-only      & ${\surd}$  &    &   & 93.33 & 94.14 \\
        inter-only     &      & ${\surd}$ &  & 93.28 & 94.46 \\
        final-only     &      &       & ${\surd}$ & 93.62 & 94.45 \\
        intra-excluded  &      & ${\surd}$ & ${\surd}$  & 93.81 & \textbf{94.72} \\
        inter-excluded & ${\surd}$ &       & ${\surd}$ & 93.48 & 94.24 \\
        final-excluded & ${\surd}$ & ${\surd}$  &  & 93.72 & 94.66 \\
        overall        & ${\surd}$ & ${\surd}$ & ${\surd}$ & \textbf{93.87} & 94.70 \\ \bottomrule
        \end{tabular}}
        \end{center}
        \caption{Ablation experiments using different parts of output features by ACAE to get context similarity.}
        \label{table:feature-ablation}
    \end{table}
    
\begin{figure}[thbp]
        \centering
        \includegraphics[width=0.75\columnwidth]{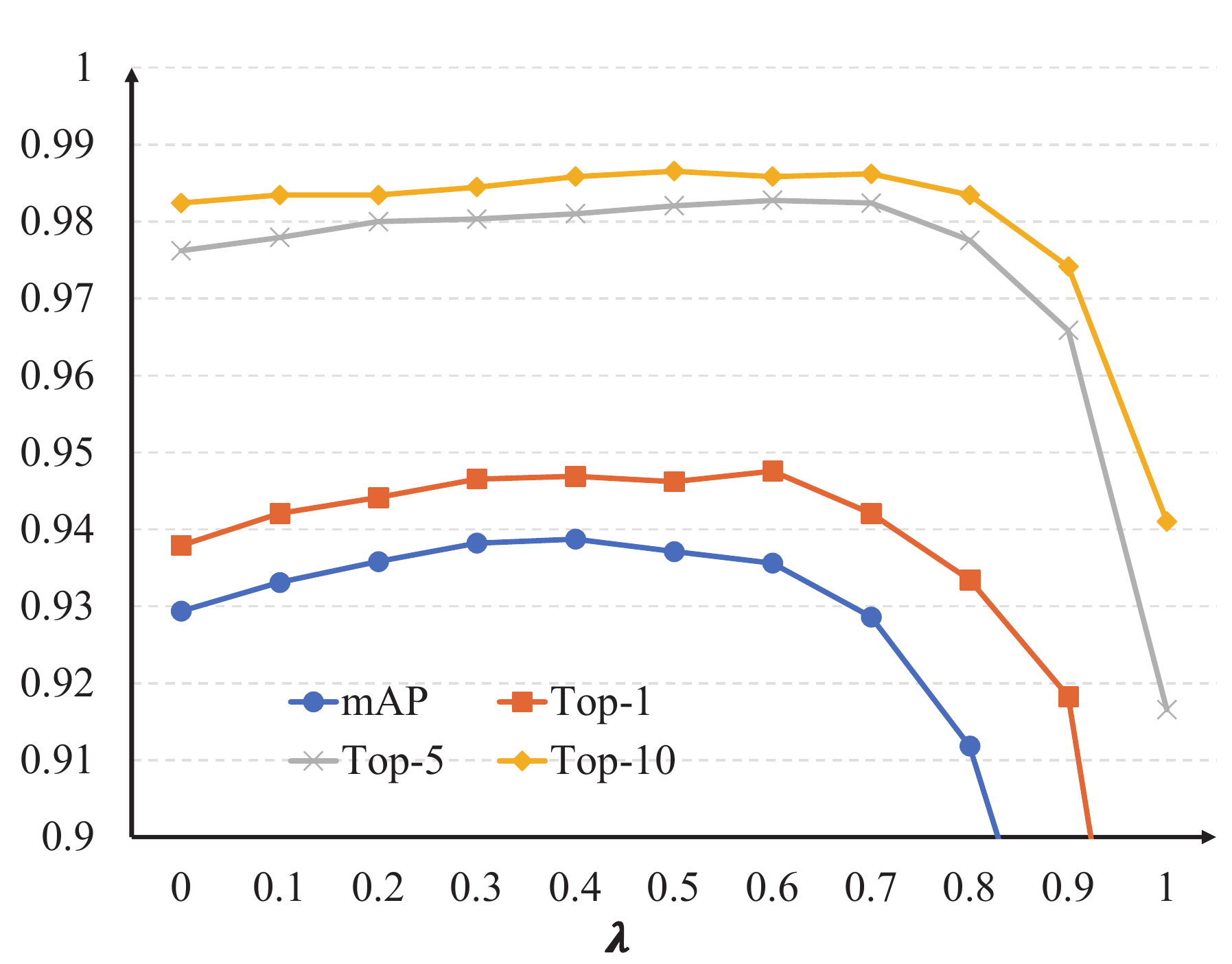}
        \caption{Evaluation performance of ACAE with varying weights of graph similarity.}
        \label{fig:exps_lambda}
    \end{figure}

    Moreover, we investigate the effect of the weight of graph similarity ${\lambda}$ in inference. Figure~\ref{fig:exps_lambda} shows that ACAE is quite robust to ${\lambda}$ since ranging ${\lambda}$ from 0.1 to 0.6 all promote the baseline performance. Best overall performance is achieved when ${\lambda}$ is set to 0.4.
    
    \textbf{How ACAE head works with different person search models.} To evaluate the expandability of ACAE, we incorporate it into two representative one-step models, OIM\cite{joint} and NAE\cite{NAE}.  Concretely, we cascade ACAE head after the individual feature branch of two models and apply IMB in training as well. We follow the same settings with these methods, except setting the batch size of NAE to 2. Results are based on CUHK-SYSU and the gallery size of 100 for a fair comparison. Table~\ref{table:ablation-acae} presents the extensive promotion of ACAE for different methods. ACAE manages to improve NAE by 0.85/0.4 mAP/Top-1 and OIM by 2.7/3.0 mAP/Top-1 respectively, as well as nearly maintaining the same performance for detection. To avoid disturbance from detectors, we further substitute detected boxes with ground-truth ones and report the promotion of ACAE in the bottom part of Table~\ref{table:ablation-acae}. Apparently, similar promotion could be also found with ACAE for all methods. Moreover, the promotion of ACAE with ground-truth boxes is almost the same as that with detected boxes, indicating that promotion of ACAE comes from the better features in re-ID instead of detection.

    \begin{table}[]
        \centering
        \setlength{\tabcolsep}{1.4mm}{
        \begin{tabular}{lccccc}
        \toprule
        \textbf{Methods}      & \textbf{Recall}   & \textbf{AP}       & \textbf{Re-ID} & \textbf{mAP} & \textbf{Top-1}  \\
        \midrule
        \multirow{2}{*}{OIM} & 81.27 & 75.77 & w/o ACAE & 76.23 & 76.28 \\
                             & 79.86 & 74.52 & w/ ACAE & 78.94 & 79.28 \\    
        \midrule
        \multirow{2}{*}{NAE} & 89.58 & 84.43 & w/o ACAE & 91.35 & 92.55 \\
                 & 89.38 & 84.46 & w/ ACAE & 92.20  & 92.86 \\
        \midrule
        \multirow{2}{*}{Ours}  & 93.25 & 90.48 & w/o ACAE & 92.59 & 93.31 \\
           & 93.67 & 91.92 & w/ ACAE & 93.87 & 94.69 \\
        \midrule
        \midrule
        \multirow{2}{*}{OIM/GT} & \multirow{2}{*}{100} & \multirow{2}{*}{100} & w/o ACAE & 79.53 & 79.00 \\
          &   &   & w/ ACAE & 81.32 & 81.21 \\ \midrule
        \multirow{2}{*}{NAE/GT}   & \multirow{2}{*}{100} & \multirow{2}{*}{100} & w/o ACAE & 92.98 & 93.59 \\
                 &       &       & w/ ACAE & 93.93 & 94.04 \\
        \midrule
        \multirow{2}{*}{Ours/GT}   & \multirow{2}{*}{100} & \multirow{2}{*}{100} & w/o ACAE & 93.61 & 94.24  \\
            &       &       & w/ ACAE & 94.78 & 95.72 \\
        \bottomrule
        \end{tabular}
        }
        \caption{Performance of ACAE on different one-step models. The bottom part shows the evaluation results with ground-truth boxes instead of detected boxes.}
        \label{table:ablation-acae}
    \end{table}
    
    \textbf{Comparison to re-ranking methods.} Re-ranking methods in Person re-ID is an effective way to promote retrieval performance, as it utilizes the 'similarity context' within gallery candidates and queries. ACAE works likewise but only concentrates on spatial contextual clues between two images, which is not always available in Person re-ID. To investigate the differences between ACAE and re-ranking methods, we provide the comparison to one of the representative re-ranking methods, k-reciprocal\cite{zhong2017re}. Moreover, we conduct full hyper-parameter searching for ${k_1, k2_, \lambda}$ based on the average performance of mAP and Top-1. Obviously, as shown in Table~\ref{table:exp-reranking}, though k-reciprocal could promote the baseline performance, ACAE achieves better performance gain both in mAP and Top-1. We believe this owes to less distraction in spatial context instead of similarity information in the whole galleries and queries.

    \begin{table}[hbtp]
        \centering
        \setlength{\tabcolsep}{1.2mm}{
            \begin{tabular}{lcccc}
            \toprule
            \textbf{Method} & \textbf{mAP} & \textbf{Top-1} & ${\Delta}$\textbf{mAP} & ${\Delta}$\textbf{Top-1} \\
            \midrule
            Ours baseline   & 92.59  & 93.31 & - & - \\
            k-reciprocal  & 92.95 & 93.72 & +0.36 & +0.41 \\
            Ours w/ ACAE   & 93.87 & 94.69 & +1.28 & +1.38 \\
            \bottomrule
            \end{tabular}
        }
        \caption{Comparison with re-ranking methods.}
        \label{table:exp-reranking}
    \end{table}
    
    \subsubsection{Visual Inspection.} 
\begin{figure*}[thbp]
        \centering
        \includegraphics[width=\textwidth]{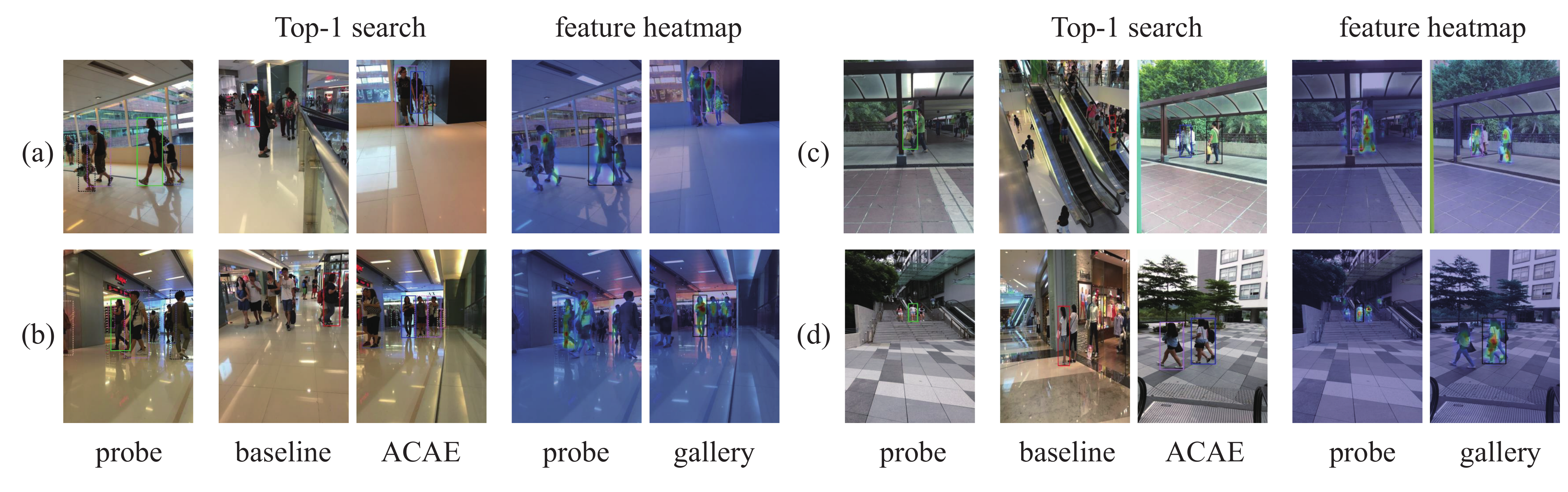}
        \caption{Visualization for several hard examples. For each query, we show the Top-1 search results of the baseline model and ACAE. Green/Red boxes denote the correct/wrong results and dashed line boxes represent the contextual persons. For the matched persons in the probe image and ACAE search result, the 'feature heatmap' part shows the discriminative regions for their corresponding contextual features. Best viewed in color.}
        \label{fig:visualization}
    \end{figure*}

To inspect whether ACAE can learn the contextual relationship of different persons across two images, we visualize some qualitative results shown in Figure~\ref{fig:visualization}. Hard cases (a)-(d) mislead the baseline model in terms of confusing individual appearance, occlusion, and large-scale variance. But, with contextual persons, ACAE successfully finds the correct matches in these cases. Besides, for the matched persons in the probe image and ACAE search result, we also apply Grad-CAM\cite{grad_cam} to interpret the discriminative regions for their corresponding final contextual features, as shown in the 'feature heatmap' part. Obviously, produced salient regions indicate that ACAE manages to focus more on possible co-travelers in feature aggregation, which corresponds to the implicit matching mechanism we desire.

    \subsection{Comparison to the State-of-the-arts}

    In this subsection, we report the comparison results of our methods with previous state-of-the-art on CUHK-SYSU and PRW. The results in Table~\ref{table:comparison-sota} show the superiority of our methods.
    
    \begin{table}
    \centering \setlength{\tabcolsep}{1mm}{
    \begin{center}
    \resizebox{\columnwidth}{!}{
    \begin{tabular}{cl|cc|cc}
    \toprule
    \multicolumn{2}{c}{\multirow{2}{*}{Methods}}& 
    \multicolumn{2}{|c|}{CUHK-SYSU} & \multicolumn{2}{c}{PRW} \\ \cline{3-6} 
    \multicolumn{2}{c|}{} & mAP & top-1 & mAP & top-1 \\ 
    \hline
    \multicolumn{1}{l|}{\multirow{6}{*}{\rotatebox{90}{two-step}}} & 
     MGTS\cite{Mask-Guided-Two-Stream} & 83.0 & 83.7 & 32.6 & 72.1 \\
    \multicolumn{1}{l|}{} & CLSA\cite{multi-scale-matching}  & 87.2 & 88.5 & 38.7 & 65.0 \\
    \multicolumn{1}{l|}{} & IGPN+PCB\cite{IGPN}  & 90.3 & 91.4 & 47.2 & 87.0 \\
    \multicolumn{1}{l|}{} & RDLR\cite{ReID-Driven}  & 93.0 & 94.2 & 42.9 & 70.2 \\
    \multicolumn{1}{l|}{} & TCTS\cite{TCTS} & 93.9 & 95.1 & 46.8 & 87.5 \\
    \hline
    \hline
    \multicolumn{1}{l|}{\multirow{10}{*}{\rotatebox{90}{one-step}}} & 
    OIM\cite{joint} & 75.5 & 78.7 & 21.3 & 49.9 \\
    \multicolumn{1}{l|}{} & IAN\cite{IAN}  & 76.3 & 80.1 & 23.0 & 61.9 \\
    \multicolumn{1}{l|}{} & NPSM\cite{NPSM} & 77.9 & 81.2 & 24.2 & 53.1 \\
    \multicolumn{1}{l|}{} & RCAA\cite{RCAA} & 79.3 & 81.3 & - & - \\
    \multicolumn{1}{l|}{} & CTXGraph\cite{CTXGraph} & 84.1 & 86.5 & 33.4 & 73.6 \\
    \multicolumn{1}{l|}{} & QEEPS\cite{QEEPS} & 88.9 & 89.1 & 37.1 & 76.7 \\
    \multicolumn{1}{l|}{} & BI-Net\cite{binet} & 90.0 & 90.7 & 45.3 & 81.7 \\
    \multicolumn{1}{l|}{} & NAE+\cite{NAE} & 92.1 & 92.9 & 44.0 & 81.1 \\
    \multicolumn{1}{l|}{} & DMR-Net\cite{decoupled} & 93.2 & 94.2 & \textbf{46.9} & 83.3 \\
    \multicolumn{1}{l|}{} & Ours ACCE & \textbf{93.9} & \textbf{94.7} & 46.2 & \textbf{86.1} \\
    \bottomrule
    \end{tabular}
    }
    \end{center}
    \caption{Comparison with previous methods on CUHK-SYSU and PRW.}
    \label{table:comparison-sota}
    }
    \end{table}
    
    \textbf{Comparison on CUHK-SYSU dataset.} In Table~\ref{table:comparison-sota}, we show the full comparison of overall performance on CUHK-SYSU between ACAE and previous state-of-the-art methods. Apparently, our overall methods outperform all previous end-to-end ones with the same ResNet50 backbone. Specifically, compared with the previous state-of-the-art DMR-Net~\cite{decoupled}, our methods surpass them by 0.7 and 0.5 in terms of mAP and top-1 respectively. Moreover, our methods achieve comparable performance with strong two-step counterpart TCTS\cite{TCTS} and surpass RDLR~\cite{ReID-Driven} by 0.9/0.5 in mAP/top-1.
    
\begin{figure}[htbp]
        \centering
        \subfigure[One-Step methods]{
            \begin{minipage}{0.45\columnwidth}
            \centering
            \includegraphics[width=\columnwidth]{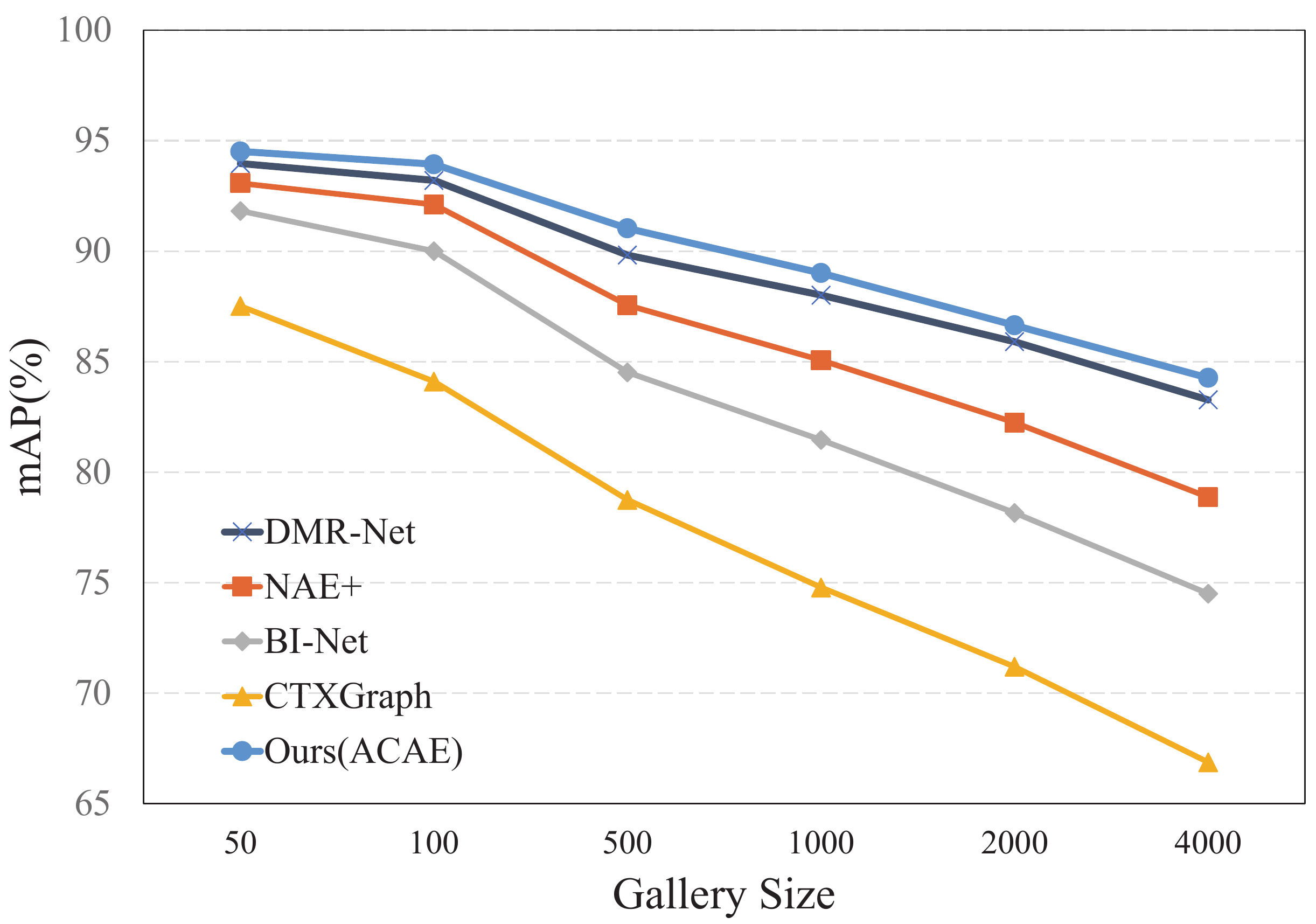}
            \label{fig:exps_gallery_size_one_step}
            \end{minipage}
        }
        \subfigure[Two-Step methods]{
            \begin{minipage}{0.45\columnwidth}
            \centering
            \includegraphics[width=\columnwidth]{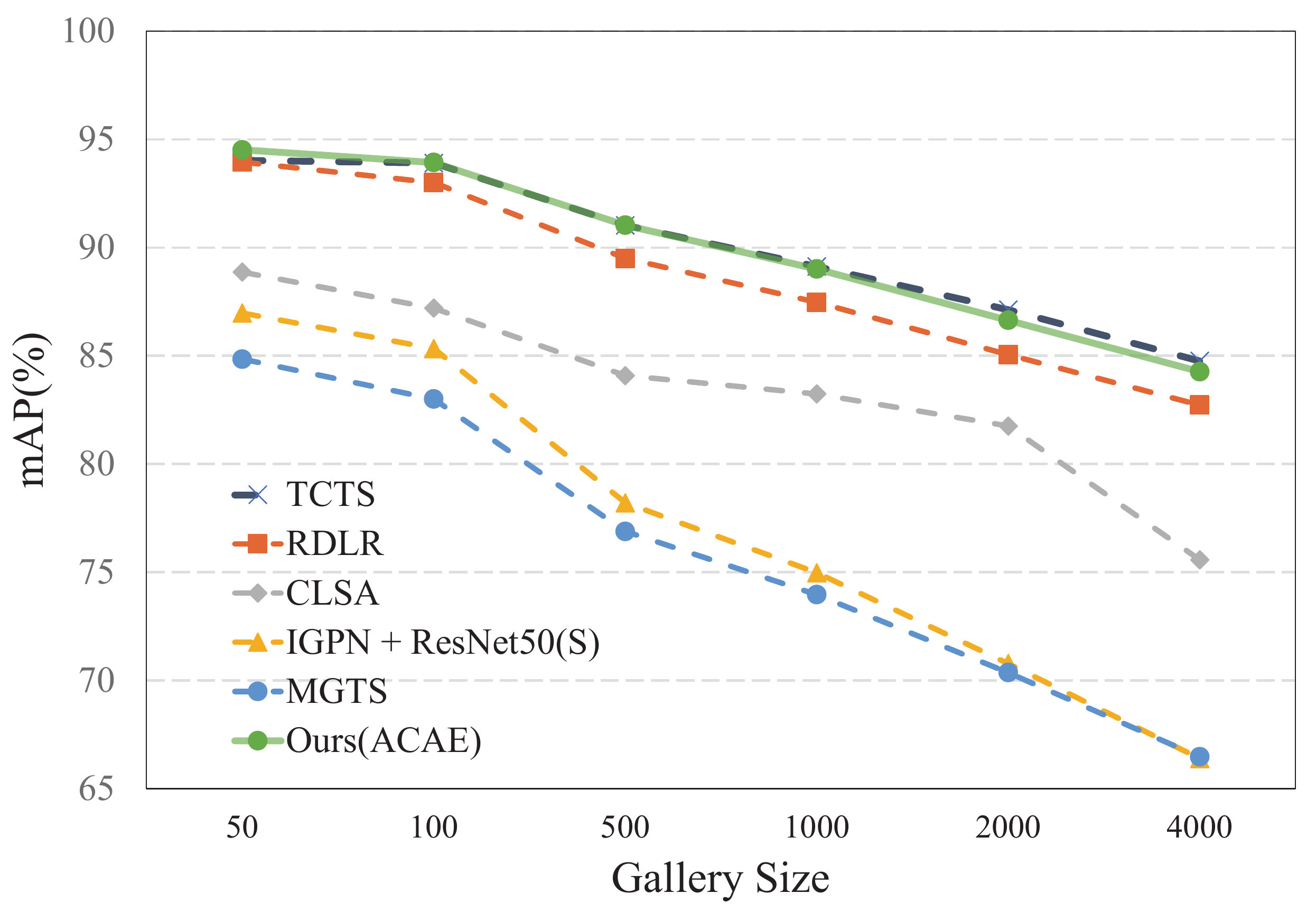}
            \label{fig:exps_gallery_size_two_step}
            \end{minipage}
        }
        \caption{Comparison with different methods with varying gallery size on CUHK-SYSU. Dashed lines present the performance of two-step methods.}
        \label{fig:exps_gallery_size}
    \end{figure}

    To verify the performance consistency of our methods, we evaluate our methods under galleries of different sizes in $[50, 100, 500, 1000, 2000, 4000]$ and report the mAP results. As the gallery size increases, more distracting persons would be introduced in the gallery, thus making searching much more challenging. However, as shown in Figure~\ref{fig:exps_gallery_size_one_step}, our methods still achieve the best performance among all previous one-step methods. Moreover, comparison in Figure~\ref{fig:exps_gallery_size_two_step} shows that ACAE is still able to obtain comparable performance with TCTS and keeps its superiority among most of the other two-step methods.
    
    \textbf{Comparison on PRW dataset.} We also evaluate the performance of our overall methods on the PRW dataset, as shown in the right column of Table~\ref{table:comparison-sota}. Compared with previous state-of-the-art methods~\cite{ReID-Driven, decoupled}, our methods achieve the best top-1 searching performance and comparable mAP with the best one-step methods and two-step methods.
    
    \textbf{Inference timing.} Additionally, we compare the average inference time of different methods with and without ACAE for a pair of images. All tests are performed with a V100 GPU for a fair comparison. As is shown in Table~\ref{table:runtime-compare}, compared with the original model, ACAE would cost roughly 5 ms on average, which is usually 3\% of total inference time. This proves that ACAE is efficient in inference.
    
    \begin{table}[t]
    \centering
    \resizebox{\columnwidth}{!}{
        \begin{tabular}{lcccc}
        \toprule
        \textbf{Method} & \textbf{Time}(ms) & \textbf{Time}(ms) w/ACAE & ${\Delta t}$ \\
        \midrule
        NAE\cite{NAE} & 124.4 & 129.3 & 4.9 \\
        Ours & 160.6 & 166.1 & 5.5 \\ 
        \bottomrule
        \end{tabular}
    }
    \caption{Runtime comparison. Time is reported as the average inference runtime of a pair of images.}
    \label{table:runtime-compare}
    \end{table}
    
\section{Conclusions}
     
    In this paper, we propose a novel feature head named \textit{Attention Context-Aware Embedding} to consider contextual information in person search. Besides, we introduce the Image Memory Bank mechanism to improve training efficiency. Extensive promotion on different models confirms the efficacy of ACAE. Based on our Transformer-based baseline model, ACAE achieves state-of-the-art on two typical benchmarks. We believe that our findings could make the community aware of the effectiveness of contextual information in person search task and draw more attention to person search. 
    
\bibliography{ref}

\begin{thebibliography}{25}
\providecommand{\natexlab}[1]{#1}

\bibitem[{Ba, Kiros, and Hinton(2016)}]{layernorm}
Ba, J.~L.; Kiros, J.~R.; and Hinton, G.~E. 2016.
\newblock Layer Normalization.
\newblock arXiv:1607.06450.

\bibitem[{Chang et~al.(2018)Chang, Huang, Shen, Liang, Yang, and
  Hauptmann}]{RCAA}
Chang, X.; Huang, P.-Y.; Shen, Y.-D.; Liang, X.; Yang, Y.; and Hauptmann, A.~G.
  2018.
\newblock Rcaa: Relational context-aware agents for person search.
\newblock In \emph{Proceedings of the European Conference on Computer Vision
  (ECCV)}, 84--100.

\bibitem[{Chen et~al.(2020{\natexlab{a}})Chen, Zhang, Ouyang, Yang, and
  Schiele}]{HOIM}
Chen, D.; Zhang, S.; Ouyang, W.; Yang, J.; and Schiele, B. 2020{\natexlab{a}}.
\newblock Hierarchical Online Instance Matching for Person Search.
\newblock In \emph{Proceedings of the AAAI Conference on Artificial
  Intelligence}, volume~34, 10518--10525.

\bibitem[{Chen et~al.(2018)Chen, Zhang, Ouyang, Yang, and
  Tai}]{Mask-Guided-Two-Stream}
Chen, D.; Zhang, S.; Ouyang, W.; Yang, J.; and Tai, Y. 2018.
\newblock Person search via a mask-guided two-stream cnn model.
\newblock In \emph{Proceedings of the European Conference on Computer Vision
  (ECCV)}, 734--750.

\bibitem[{Chen et~al.(2020{\natexlab{b}})Chen, Zhang, Yang, and Schiele}]{NAE}
Chen, D.; Zhang, S.; Yang, J.; and Schiele, B. 2020{\natexlab{b}}.
\newblock Norm-aware embedding for efficient person search.
\newblock In \emph{Proceedings of the IEEE/CVF Conference on Computer Vision
  and Pattern Recognition}, 12615--12624.

\bibitem[{Dong et~al.(2020{\natexlab{a}})Dong, Zhang, Song, and Tan}]{binet}
Dong, W.; Zhang, Z.; Song, C.; and Tan, T. 2020{\natexlab{a}}.
\newblock Bi-directional interaction network for person search.
\newblock In \emph{Proceedings of the IEEE/CVF Conference on Computer Vision
  and Pattern Recognition}, 2839--2848.

\bibitem[{Dong et~al.(2020{\natexlab{b}})Dong, Zhang, Song, and Tan}]{IGPN}
Dong, W.; Zhang, Z.; Song, C.; and Tan, T. 2020{\natexlab{b}}.
\newblock Instance guided proposal network for person search.
\newblock In \emph{Proceedings of the IEEE/CVF Conference on Computer Vision
  and Pattern Recognition}, 2585--2594.

\bibitem[{Han et~al.(2019)Han, Ye, Zhong, Tan, Zhang, Gao, and
  Sang}]{ReID-Driven}
Han, C.; Ye, J.; Zhong, Y.; Tan, X.; Zhang, C.; Gao, C.; and Sang, N. 2019.
\newblock Re-id driven localization refinement for person search.
\newblock In \emph{Proceedings of the IEEE/CVF International Conference on
  Computer Vision}, 9814--9823.

\bibitem[{Han et~al.(2021)Han, Zheng, Gao, Sang, and Yang}]{decoupled}
Han, C.; Zheng, Z.; Gao, C.; Sang, N.; and Yang, Y. 2021.
\newblock Decoupled and Memory-Reinforced Networks: Towards Effective Feature
  Learning for One-Step Person Search.

\bibitem[{He et~al.(2017)He, Gkioxari, Doll{\'a}r, and Girshick}]{mask-rcnn}
He, K.; Gkioxari, G.; Doll{\'a}r, P.; and Girshick, R. 2017.
\newblock Mask r-cnn.
\newblock In \emph{Proceedings of the IEEE international conference on computer
  vision}, 2961--2969.

\bibitem[{Lan, Zhu, and Gong(2018)}]{multi-scale-matching}
Lan, X.; Zhu, X.; and Gong, S. 2018.
\newblock Person search by multi-scale matching.
\newblock In \emph{Proceedings of the European conference on computer vision
  (ECCV)}, 536--552.

\bibitem[{Liu et~al.(2017)Liu, Feng, Jie, Jayashree, Zhao, Qi, Jiang, and
  Yan}]{NPSM}
Liu, H.; Feng, J.; Jie, Z.; Jayashree, K.; Zhao, B.; Qi, M.; Jiang, J.; and
  Yan, S. 2017.
\newblock Neural person search machines.
\newblock In \emph{Proceedings of the IEEE International Conference on Computer
  Vision}, 493--501.

\bibitem[{Munjal et~al.(2019)Munjal, Amin, Tombari, and Galasso}]{QEEPS}
Munjal, B.; Amin, S.; Tombari, F.; and Galasso, F. 2019.
\newblock Query-guided end-to-end person search.
\newblock In \emph{Proceedings of the IEEE/CVF Conference on Computer Vision
  and Pattern Recognition}, 811--820.

\bibitem[{Ren et~al.(2015)Ren, He, Girshick, and Sun}]{faster-rcnn}
Ren, S.; He, K.; Girshick, R.~B.; and Sun, J. 2015.
\newblock Faster {R-CNN:} Towards Real-Time Object Detection with Region
  Proposal Networks.
\newblock In \emph{Advances in Neural Information Processing Systems 28: Annual
  Conference on Neural Information Processing Systems 2015, December 7-12,
  2015, Montreal, Quebec, Canada}, 91--99.

\bibitem[{Selvaraju et~al.(2017)Selvaraju, Cogswell, Das, Vedantam, Parikh, and
  Batra}]{grad_cam}
Selvaraju, R.~R.; Cogswell, M.; Das, A.; Vedantam, R.; Parikh, D.; and Batra,
  D. 2017.
\newblock Grad-cam: Visual explanations from deep networks via gradient-based
  localization.
\newblock In \emph{Proceedings of the IEEE international conference on computer
  vision}, 618--626.

\bibitem[{Vaswani et~al.(2017)Vaswani, Shazeer, Parmar, Uszkoreit, Jones,
  Gomez, Kaiser, and Polosukhin}]{transformer}
Vaswani, A.; Shazeer, N.; Parmar, N.; Uszkoreit, J.; Jones, L.; Gomez, A.~N.;
  Kaiser, {\L}.; and Polosukhin, I. 2017.
\newblock Attention is all you need.
\newblock In \emph{Advances in neural information processing systems},
  5998--6008.

\bibitem[{Veli{\v{c}}kovi{\'c} et~al.(2018)Veli{\v{c}}kovi{\'c}, Cucurull,
  Casanova, Romero, Li{\`o}, and Bengio}]{velivckovic2018GAT}
Veli{\v{c}}kovi{\'c}, P.; Cucurull, G.; Casanova, A.; Romero, A.; Li{\`o}, P.;
  and Bengio, Y. 2018.
\newblock Graph Attention Networks.
\newblock In \emph{International Conference on Learning Representations}.

\bibitem[{Wang et~al.(2020)Wang, Ma, Chang, Shan, and Chen}]{TCTS}
Wang, C.; Ma, B.; Chang, H.; Shan, S.; and Chen, X. 2020.
\newblock Tcts: A task-consistent two-stage framework for person search.
\newblock In \emph{Proceedings of the IEEE/CVF Conference on Computer Vision
  and Pattern Recognition}, 11952--11961.

\bibitem[{Xiao et~al.(2019)Xiao, Xie, Tillo, Huang, Wei, and Feng}]{IAN}
Xiao, J.; Xie, Y.; Tillo, T.; Huang, K.; Wei, Y.; and Feng, J. 2019.
\newblock IAN: the individual aggregation network for person search.
\newblock \emph{Pattern Recognition}, 87: 332--340.

\bibitem[{Xiao et~al.(2017)Xiao, Li, Wang, Lin, and Wang}]{joint}
Xiao, T.; Li, S.; Wang, B.; Lin, L.; and Wang, X. 2017.
\newblock Joint detection and identification feature learning for person
  search.
\newblock In \emph{Proceedings of the IEEE Conference on Computer Vision and
  Pattern Recognition}, 3415--3424.

\bibitem[{Xu et~al.(2014)Xu, Ma, Huang, and Lin}]{commonness}
Xu, Y.; Ma, B.; Huang, R.; and Lin, L. 2014.
\newblock Person search in a scene by jointly modeling people commonness and
  person uniqueness.
\newblock In \emph{Proceedings of the 22nd ACM international conference on
  Multimedia}, 937--940.

\bibitem[{Yan et~al.(2019)Yan, Zhang, Ni, Zhang, Xu, and Yang}]{CTXGraph}
Yan, Y.; Zhang, Q.; Ni, B.; Zhang, W.; Xu, M.; and Yang, X. 2019.
\newblock Learning context graph for person search.
\newblock In \emph{Proceedings of the IEEE/CVF Conference on Computer Vision
  and Pattern Recognition}, 2158--2167.

\bibitem[{Zheng et~al.(2017)Zheng, Zhang, Sun, Chandraker, Yang, and
  Tian}]{PRW}
Zheng, L.; Zhang, H.; Sun, S.; Chandraker, M.; Yang, Y.; and Tian, Q. 2017.
\newblock Person re-identification in the wild.
\newblock In \emph{Proceedings of the IEEE Conference on Computer Vision and
  Pattern Recognition}, 1367--1376.

\bibitem[{Zhong et~al.(2017)Zhong, Zheng, Cao, and Li}]{zhong2017re}
Zhong, Z.; Zheng, L.; Cao, D.; and Li, S. 2017.
\newblock Re-ranking person re-identification with k-reciprocal encoding.
\newblock In \emph{Proceedings of the IEEE Conference on Computer Vision and
  Pattern Recognition}, 1318--1327.

\bibitem[{Zhu et~al.(2020)Zhu, Su, Lu, Li, Wang, and Dai}]{deformable-detr}
Zhu, X.; Su, W.; Lu, L.; Li, B.; Wang, X.; and Dai, J. 2020.
\newblock Deformable DETR: Deformable Transformers for End-to-End Object
  Detection.

\end{thebibliography}

\end{document}